\documentclass[letterpaper]{article} % DO NOT CHANGE THIS
\usepackage{aaai23}  % DO NOT CHANGE THIS
\usepackage{times}  % DO NOT CHANGE THIS
\usepackage{helvet}  % DO NOT CHANGE THIS
\usepackage{courier}  % DO NOT CHANGE THIS
\usepackage[hyphens]{url}  % DO NOT CHANGE THIS
\usepackage{graphicx} % DO NOT CHANGE THIS
\usepackage{amssymb} % Added for \mathbb support
\usepackage{natbib}  % DO NOT CHANGE THIS AND DO NOT ADD ANY OPTIONS TO IT
\usepackage{caption} % DO NOT CHANGE THIS AND DO NOT ADD ANY OPTIONS TO IT
\usepackage{algorithm}
\usepackage{algorithmic}

\usepackage{newfloat}
\usepackage{listings}
\DeclareCaptionStyle{ruled}{labelfont=normalfont,labelsep=colon,strut=off} % DO NOT CHANGE THIS
\floatstyle{ruled}
\newfloat{listing}{tb}{lst}{}
\floatname{listing}{Listing}
\title{Application of LLMs to Multi-Robot Path Planning and Task Allocation}
\author{
    Ashish Kumar
}
\affiliations{
    College of Engineering, Northeastern University\\
    Boston, Massachusetts 02115 USA\\
    kumar.ashish1@northeastern.edu
}

\begin{document}

\maketitle

\begin{abstract}
Efficient exploration is a well known problem in deep reinforcement learning and this problem is exacerbated in multi-agent reinforcement learning due the intrinsic complexities of such algorithms. There are several approaches to efficiently explore an environment to learn to solve tasks by multi-agent operating in that environment, of which, the idea of expert exploration is investigated in this work. More specifically, this work investigates the application of large-language models as expert planners for efficient exploration in planning based tasks for multiple agents.
\end{abstract}

\section{Introduction}
The work aims to address the challenges of exploration for reinforcement learning-based agents for Multi-Agent Path Finding (MAPF) problems. MAPF is a critical component in various applications, including robotics, logistics, and autonomous vehicle navigation, where multiple agents must find optimal paths from their respective starting points to their destinations without collisions. Most times, agents are not just navigating in an environment, they also perform tasks so path planning and multi-robot task allocation (MRTA) go hand in hand. There are various approaches to solve MAPF and MRTA at scale, of which, reinforcement learning-based, specifically, multi-agent reinforcement learning (MARL) based approaches have shown promising results in recent times\cite{Sartoretti_2019}  \cite{Damani_2021}.

Despite their successes, deep reinforcement learning (DRL) and deep MARL are limited by significant sample inefficiency, requiring millions of interactions for even relatively simple tasks, which hampers broader real-world application. As mentioned in the previous paragraph, effectively navigating an environment to gather useful data can enhance policy learning toward optimal solutions. This challenge is exacerbated in complex environments characterized by sparse rewards, noise, long decision horizons, and nonstationary agents \cite{10021988}. 

The idea of utilizing the knowledge of a task expert in reinforcement learning is not new and has been explored in both online and offline reinforcement learning settings \cite{9694460} and \cite{kumar2022offline}. The intuition is to enable an agent to explore more rewarding or optimal trajectories in the state, action space as opposed to a less focused exploration allowing the agent to learn to assign high value to such joint sub-spaces. 

The hypothesis for this work is that large-language models are good or near-optimal planners in simple path and action planning tasks. Furthermore, such LLMs can be applied as a method for expert exploration in MARL.

\section{Related Work}
The following is a compilation of scholarly articles/papers that have been reviewed for this work:
\begin{enumerate}
    \item \textit{Guiding Pretraining in Reinforcement Learning with Large Language Models} - This paper is an inspiration for this work. The paper introduces ELLM, an intrinsic motivation method that uses pre-trained Large Language Models (LLMs) to guide exploration in reinforcement learning toward behaviors grounded in common sense.\cite{du2023guiding}
    \item \textit{PlanBench} - The paper introduces PlanBench, a benchmark suite designed to evaluate large language models (LLMs) on their ability to reason and plan within various domains. Despite testing on relatively straightforward scenarios, the findings reveal that LLMs often underperform in tasks requiring common-sense planning and reasoning about actions and changes. This paper helped me understand the application of LLMs to planning and reasoning problems and inherent complexities.\cite{valmeekam2023planbench}
    \item \textit{Can LLMs be good path planners?} - This paper examines large language models' (LLMs) capacity for spatial-temporal reasoning through a new dataset named PPNL, focused on path-planning tasks. The research demonstrates LLMs' strengths in spatial reasoning when provided with specific spatial information and continuous environmental feedback. The dataset from this work will be used for fine-tuning an LLM, which is to be determined.\cite{aghzal2024large}
    \item \textit{Large Language Models for Robotics} - This paper reviews various applications of LLMs in robotics. The results from the experiments described in this work show that LLMs demonstrate impressive reasoning, language understanding, and multimodal processing abilities that can significantly enhance robots’ comprehension of instructions, environments, and required actions. \cite{wang2024large}
    \item \textit{Enabling Intelligent Interactions between an Agent and an LLM: A Reinforcement Learning Approach} - This paper introduces "When2Ask," a novel reinforcement learning (RL) strategy designed to optimize the cost-effectiveness of interactions between local agents and large language models (LLMs). Under this approach, a "mediator" is trained as an explicit asking policy using RL to learn when to interact with a planner which is Vicuna-7B based \cite{hu2024enabling}.
determine when to interact with the planner.
    \item \textit{Judging LLM-as-a-Judge with MT-Bench and Chatbot Arena} - This paper introduced the Vicuna-7B LLM and introduced the concept of LLM-as-a-judge for chatbot evaluation \cite{zheng2023judging}.
\end{enumerate}

\section{Background}
This section provides an overview of the basics of reinforcement learning, deep reinforcement learning, multi-agent reinforcement learning, and large-language models. To begin with, Reinforcement learning is a type of machine learning where an agent learns to make decisions by performing actions in an environment and receiving feedback in the form of rewards or penalties. Reinforcement learning is widely used in various applications, such as game playing, robotic control, and autonomous vehicle navigation, and is usually solved by formulating a system as a Markov Decision Process. Markov Decision Process or finite MDP – A mathematical framework used for modeling decision making in a dynamic system governed by a probabilistic dynamics and defines an environment for reinforcement learning. It is defined as a tuple comprising of seven elements -- ($\mathcal{S}$, $\mathcal{A}$, $\mathcal{T}$, \textit{r}, $\gamma$, $\mathcal{S}_0$, $\mathcal{H}$), where $\mathcal{S}$ is the state space, $\mathcal{A}$ is the action space, $\mathcal{T}$ is the state transition probability function $\mathcal{T}$ = $\mathcal{P}({s_{t+1}}|{s_t}, {a_t})$, \textit{r} is the environment reward function \textit{r}: $\mathcal{S}\times\mathcal{A}\rightarrow\mathbb{R}^{1}$, $\gamma$ is the discount value, $\mathcal{S}_0$ is the start state distribution and $\mathcal{H}$ is the horizon length. Return or discounted return is defined as the sum of discounted rewards from current time to end of horizon T. 
In reinforcement learning, two functions estimate the value of a state or a state and action pair. They are the \emph{state value} function and the {\emph{state-action value} or \emph{Q-function} defined in Equations \ref{eq:value-function} and \ref{eq:q-function} respectively:
    \begin{equation}
        \label{eq:value-function}
        \mathcal{V}^\pi(s) = \mathbb{E}[\sum\limits_{t=0}^H\gamma^tr_t \mid s_0=s]
    \end{equation}
    
    \begin{equation}
        \label{eq:q-function}
        \mathcal{Q}^\pi(s,a) = \mathbb{E}[\sum\limits_{t=0}^H\gamma^tr_t \mid s_0=s, a_0=a]
    \end{equation}

Depending on the distribution of action used for collecting data for learning, reinforcement learning has two categories, viz. \textit{on-policy} and \textit{off-policy}. In an on-policy setting, the algorithm generates an action for a particular state based on the current policy, whereas, in the case of off-policy, the data used for learning can be from a previous version of the current policy or from an entirely different distribution. Q-learning is a well-known off-policy learning method for which the objective is to minimize the Bellman error, defined as:
    \begin{equation}
        \label{eq:q-learning-objective-function}
        \mathcal{L}(\theta) = (\mathcal{Q}_\theta(s_t, a_t) - (r_t + \gamma\max_{a}\mathcal{Q}(s_{t+1}, a)))
    \end{equation} 

Deep reinforcement learning combines the principles of reinforcement learning (RL) with deep learning. In this approach, deep neural networks are used to approximate the optimal policy and/or value functions involved in decision-making tasks. These networks enable the agent to handle high-dimensional input spaces, such as raw pixel data from video frames, which are common in complex environments. To improve efficiency and stability, deep RL often uses past experiences (stored in memory) to train the network, breaking correlations between sequential observations.

Uncertainty estimation in reinforcement learning involves quantifying the confidence of a model in its learned behaviors and predictions. These concepts are crucial for making informed decisions in environments where the outcomes are uncertain or the data is incomplete. Uncertainty can arise from various sources, including the inherent stochasticity of the environment, incomplete knowledge about the environment's dynamics, or epistemic uncertainty of an algorithm. In the context of online reinforcement learning, uncertainty estimation has been used for efficient exploration strategies such as upper confidence bound exploration via Q-ensembles \cite{chen2017ucb}.

Multi-agent reinforcement learning (MARL) is a branch of reinforcement learning where multiple agents learn simultaneously within a shared environment. Each agent makes decisions based on its own observations and possibly the inferred or observed actions of other agents. The key challenge in MARL is that the environment becomes non-stationary from the perspective of any single agent because the behavior of other agents changes as they learn, which can affect the outcomes of actions unpredictably \cite{marl-book}. There are various problem-specific formulations of a system for MARL of which Dec-POMDP or decentralized partially observable Markov decision process is a well-known formulation. It is defined as a tuple comprising of seven elements -- $(S, \{A_i\}), T, R, \{\Omega_i\}, O, \gamma)$ where $S$ is the state space, $A_i$ is a set of actions for agent $i$, with $A=\times_i A_i$ is the set of joint actions, $T$ is the state transition probability function, $\Omega_i$ is a set of observations for agent $i$, $O$ is the observation probabilities, and $\gamma$ is the discount value.

The QMIX paper \cite{rashid2018qmix}, titled "QMIX: Monotonic Value Function Factorisation for Deep Multi-Agent Reinforcement Learning," presents a novel approach for handling cooperative multi-agent scenarios. It introduces QMIX, an algorithm that efficiently decomposes the joint action-value function into a set of per-agent value functions that are combined via a learnable mixing network. This mixing network is structured to ensure that the overall action-value function remains monotonic concerning the per-agent contributions, which guarantees consistency between the joint action-value function and individual agent contributions.

Language models are computational tools designed to understand, generate, and manipulate human language. These models predict the likelihood of a sequence of words, thereby enabling a wide range of applications such as text completion, translation, and summarization. Language models are trained on large datasets of text and learn the structure and nuances of language from this data. Causal language models are a subset of language models that generate text sequentially, predicting the next word in a sequence given all the previous words. This type of model operates under the principle of causality, meaning that each word generated is causally influenced by the words that precede it. Large language models are advanced language models designed to process, understand, and generate human language. They are built using deep learning techniques, particularly transformer architectures, which allow them to handle extensive amounts of data and complex patterns in language. These models are trained on diverse and massive text corpora to capture a wide range of linguistic nuances and contexts, and demonstrate advanced properties like in-context learning. 

\section{Description}
\label{description}
In this work, I am proposing a novel approach to introduce expert exploration when training a MARL algorithm in a grid-like or 2D environment with multiple agents cooperatively solving a task. The expert in this work is either an A* algorithm, which is a well-known pathfinding algorithm, or a large-language model (LLM) based planner. This system has been designed in a way that an expert planner assists a multi-agent reinforcement learning algorithm (MARL) with better exploration when the algorithm has high uncertainty. The algorithm's intrinsic uncertainty is estimated with the help of an ensemble of mixer networks. Given a $(s, a)$, state-action pair, an ensemble of mixer networks: $\{Q_{{mix}_i}\ \forall i \in I\}$, where $I$ is the set of all agents, estimates q-value for $(s, a)$. The standard deviation of the q-values is used as an indicator of the algorithm's intrinsic uncertainty, where, higher standard deviation indicates higher model uncertainty. Below is a system diagram for a better understanding of the system,

\begin{figure}[!htb]
    \centering
    \includegraphics[width=1\linewidth]{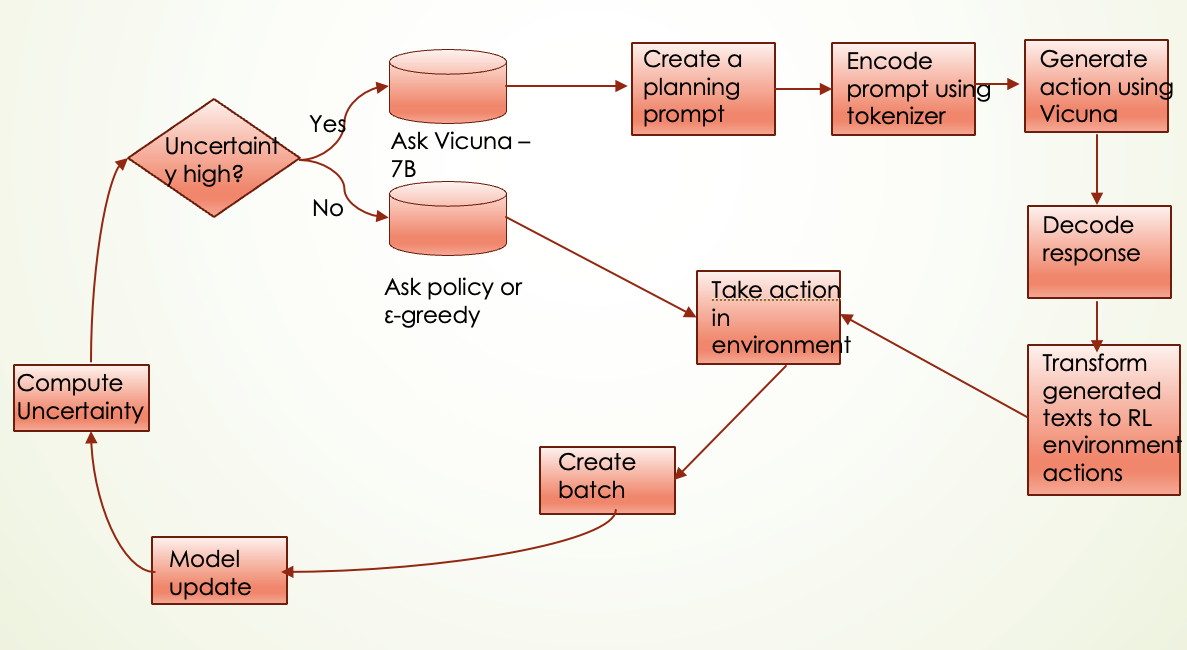}
    \caption{System Diagram}
    \label{fig:enter-label}
\end{figure}

As mentioned above, the uncertainty for the QMIX algorithm is determined using \textit{n} critics, where \textit{n} is configurable. The q-mixers are trained separately and predict the values for a state. If the standard deviation of the q-estimates is higher than a threshold then Vicuna is asked for a plan. A code snippet from the work is shown below:
\begin{lstlisting}
self._use_ensemble = ensemble_size > 1
if self._use_ensemble:
  self._q_mixer_ensemble = [QMixingNetwork(**mixing_net_kwargs) for _ in     range(ensemble_size)]
  self._q_mixer_ensemble_optimizers = [Adam(self._q_mixer_ensemble[i].parameters(), lr=self._critic_lr) for i in range(ensemble_size)]
  self._uncertainty_threshold = uncertainty_threshold
----
if self._use_ensemble:
  q_values = torch.stack([model(observation, agent_ids) for model in self._q_mixer_ensemble] + [self._q_mixer(observation, agent_ids)])
  q_values_mean = q_values.mean(dim=0)
  q_values_std = q_values.std(dim=0)
  if q_values_std.mean() > self._uncertainty_threshold:
    actions = self._ask_vicuna_for_action(observation)
\end{lstlisting}

For an LLM-based expert planner, the choice of Vicuna-7B is inspired by the work from \cite{hu2024enabling}. To integrate Vicuna-7B as a planner in the system, two major steps are performed, viz. initialization of the Vicuna-7B model and tokenizer using Huggingface APIs provided in an open-source library available here: https://github.com/lm-sys/FastChat, which is an implementation provided by \cite{hu2024enabling}. Secondly, to query Vicuna-7B, this work creates a planning prompt template specifically for planning in the SimpleSpread environment provided under PettingZoo environments \cite{mordatch2017emergence}, \cite{lowe2017multi}. The template is shown below:
\begin{lstlisting}
prompt = f"There are {self._num_agents} agents in the environment. The agents are working in a grid world and all agents are globally rewarded based on how far the closest agent is to each landmark. " \
f"Locally, the agents are penalized if they collide with other agents. The possible actions are: 0: nothing, 1: left, 2: right, 3: down, and 4: up. " \
f"Please help the agents to plan the next actions given agents' current observations. The actions should be displayed in a list. Do not explain the reasoning. " \
f"The first agent is at position {[round(x, 2) for x in locations[0:2]]}, the closest landmarks are at {[round(x, 2) for x in locations[6:]]}. " \
f"The second agent is at position {[round(x, 2) for x in locations[2:4]]}, the closest landmarks are at {[round(x, 2) for x in locations[6:]]}. " \
f"The third agent is at position {[round(x, 2) for x in locations[4:6]]}, the closest landmarks are at {[round(x, 2) for x in locations[6:]]}. " \
f"What are the next actions for the agents? The output should be a list of integers with length {self._num_agents}."
\end{lstlisting}

For fine-tuning the Vicuna-7B, there are two parameters to be used in a config file corresponding to an experiment, viz. \textit{fine\_tune\_vicuna} and \textit{fine\_tune\_samples}. \textit{fine\_tune\_vicuna} accepts a boolean value and \textit{fine\_tune\_samples} accepts a numeric value representing the number of individual data samples to use for fine-tuning. Let's say that $F$ is the prompt function, $o$ is a combined observation from the environment for all agents, $a$ is the combined action from an expert for all the agents, then the dataset for fine-tuning is described as $\{(F(o_t), a_t) \forall t \in T \}$, where $T$ is the total timesteps that agents interact with the environment using "expert" planner.

\section{Experiments}

The experiments conducted for this work contain a systematic exploration of various configurations combining QMIX,  with different enhancements such as attention mechanisms, expert systems like A*, and a presumably more advanced multi-agent planner, Vicuna-7B. Here's a breakdown of each experiment:

\begin{enumerate}
    \item \textit{Vanilla QMIX with RNN} - This experiment uses the standard QMIX architecture with an RNN Layer, which is designed to enable centralized training with decentralized execution for MARL. However, there is a slight in that the GRU layer used in the original layer has been replaced with an LSTM layer.
    
    \item \textit{QMIX with Attention layer replacing RNN} - Replacing the RNN with an attention layer shifts how the model processes temporal and spatial dependencies among agents. Attention layers can provide a better way of aggregating information across different agents and time, by avoiding or minimizing the vanishing gradient issues arising from Back-propagation through time (BPTT). This could lead to more efficient learning in environments where the relationships between agents' observations are complex and dynamically changing. This is not a novel idea and the application of transformers have been explored in the work by \cite{gallici2023transfqmix}.
    \item \textit{QMIX + Attention Layer using A* as oracle} - In this experiment, one of the core idea proposed in this work has been tested. As mentioned in \ref{description}, in this experiment an A* star based single-agent pathfinding algorithm is used to generate "expert" plans for individual agents, one at a time and the planner does not make any "global" decisions. 
    \item \textit{QMIX + Attention + Vicuna-7B} - In this experiment, the main idea proposed in this work has been tested. The A* is replaced by the Vicuna-7B LLM as the "expert" multi-agent planner. The invocation of the LLM to plan actions for the agent is performed in the same way as described above. The system diagram provided in \ref{description} provides an illustration of this process. No fine tuning is performed in this section.
    \item \textit{QMIX + Attention + Vicuna-7B fine tuned} - As the name suggests, the Vicuna-7B is fined tuned in this experiments using "supervisory" dataset prepared using the A* algorithm. Prior to training the QMIX algorithm, the Vicuna-7B is fine-tuned.
\end{enumerate}

\subsubsection{Specification}

The experimental evaluations were conducted using the Simple Spread environment from the \textit{PettingZoo} library, a popular framework for multi-agent reinforcement learning research. The computational setup consisted of a local computer running Ubuntu 22.04.4 LTS. The hardware configuration included an Intel Core i7-11700 processor, an NVIDIA GeForce GTX 1070 Ti graphics processing unit, and 16 GB of system RAM. Additionally, for fine tuning the Vicuna-7B, the NVIDIA A100 Tensor Core GPU available via the Google Colab platform was utilized.

\subsection{Result}
\begin{enumerate}
    \item \textit{Vanilla QMIX with RNN} - In this experiment, the agent's performance during the evaluation rounds post each iteration was oscillating and did not stabilize during the later stages of the training. The performance of the agent is lower compared to the other experiments. My hypothesis is that the LSTM models are harder to train for QMIX, hence the authors chose GRU in the QMIX paper. 
\begin{figure}[!htb]
        \centering
        \includegraphics[width=0.5\linewidth]{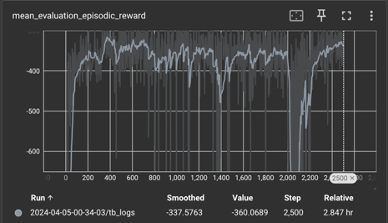}
        \caption{Vanilla QMIX with RNN}
        \label{fig:enter-label}
    \end{figure}
        
    \item \textit{QMIX with Attention layer replacing RNN} - Result from this experiment shows improvement in terms of performance and stability of the agent as compared to the vanilla QMIX. This results supports the hypothesis described in the previous point and in the beginning of this section. 

\begin{figure}[!htb]
    \centering
    \includegraphics[width=0.5\linewidth]{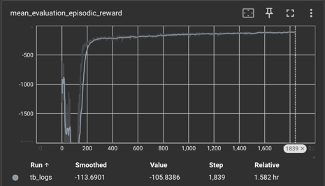}
    \caption{QMIX with Attention layer}
    \label{fig:enter-label}
\end{figure}
    
    \item \textit{QMIX + Attention Layer using A* as oracle} - The results in this experiment validate one of the main hypothesis of the paper which is that "expert" based exploration improves the performance of a MARL algorithm. Although the performance is less stable which might be attributed to A* not being a "multi-agent" algorithm, but this phenomenon requires further studies.

    \begin{figure}[!htb]
        \centering
        \includegraphics[width=0.5\linewidth]{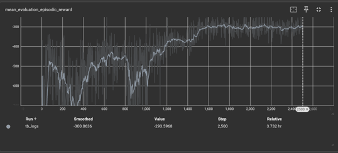}
        \caption{QMIX + Attention Layer using A*}
        \label{fig:enter-label}
    \end{figure}
    \item \textit{QMIX + Attention + Vicuna-7B} - The results in this experiment validate \textit{the} main hypothesis of the paper which is that LLM can be "expert" planners for multiple agents performing a collaborative task in a grid world like environment. The agent outperforms the agent trained in the other experiments and performance is more stable that the previous experiment suggesting that Vicuna-7B is a better planner than A* in a multi-agent setting with a caveat that this hypothesis needs to be confirmed in more complex and higher-dimensional setting.

    \begin{figure}
        \centering
        \includegraphics[width=0.5\linewidth]{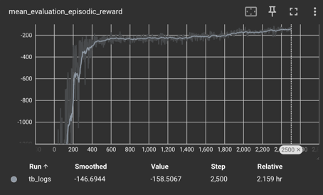}
        \caption{QMIX + Attention + Vicuna-7B}
        \label{fig:enter-label}
    \end{figure}
    \item \textit{QMIX + Attention + Vicuna-7B fine tuned} - The results of this experiment were similar to the previous based on visual inspection, however, the result could not be retrieved due to a bug in the source code for experiment naming. My hypothesis is that the results should improve with more fine-tuning as the fine-tuning was performed using 1000 "planning" conversations. Additionally, the pre-trained Vicuna-7B outputted invalid actions occasionally.
\end{enumerate}

\section{Broader Implications}

Exploration in multi-agent reinforcement learning (MARL) poses unique challenges, especially when dealing with off-policy learning. Off-policy learning, where the policy being learned is different from the policy used to generate data, allows agents to learn from past experience or from experiences generated by other policies. This approach is particularly useful in multi-agent settings where interactions are complex and environments are dynamic, combined with proven results in the such problem setting. However, the problem of exploration becomes more pronounced due to the need to efficiently learn useful policies while managing interactions between agents \cite{pmlr-v139-liu21j}, \cite{10021988}.

To address these challenges, some direction might be to focus on developing more sophisticated exploration strategies that can account for the actions and learning processes of other agents, perhaps by integrating concepts from game theory and mechanism design. Furthermore, advancements in understanding and modeling the multi-agent dynamics can lead to better off-policy correction techniques that mitigate the issues of non-stationarity and historical data reliance.

Expert exploration provides a form of informed guidance that allows agents to focus on relevant parts of the state or action space, reducing the time and experience required to learn effective policies. By following expert demonstrations or advice, agents can quickly acquire complex behaviors that might take much longer to learn through random exploration. 

Although, LLMs are large and general-purpose models, they are over-complex and slower for the purposes of planning on robots as compared to MARL based algorithms, which are more specialized for sequential decision making and may outperform LLMs with sufficient and proper training. Due to potential of application of MARL in areas like Multi-Robot Warehouse Management, Competitive Play in Board Games and Video Games, Autonomous Driving and Automated Trading in Electronic Markets as described in \cite{marl-book}, approaches that may improve the performance of off-policy MARL algorithms may have broader implications.

\section{Conclusion}
This work provides a novel approach to introduce Vicuna-7B based expert exploration when training a QMIX agent. This work validates that LLM based planners are a good candidate to be selected as an expert in an expert based exploration approach in MARL. Since, Vicuna-7B is a smaller model compared to the kinds of Llama-2, I hypothesize that the larger model may even lead to faster convergence of the QMIX algorithm and even further improve the performance of QMIX in an environment. Based on empirical data and qualitative analysis, the actions produced by Vicuna-7B were completely bad in some cases. This is where, I hypothesize that larger models may be better and this has been identified as a direction for future work mentioned below. 

\section{Future Directions}

For future work, I have identified the following areas:
\begin{enumerate}
    \item \textit{Tune the number of ensemble agents parameter} - Adjusting the number of ensemble agents can help improve the robustness and accuracy of uncertainty estimation as the uncertainty estimation is based on the standard deviation of the q-estimates performed by the ensemble network.
\item \textit{Investigate a better epistemic uncertainty estimation approach} - Epistemic uncertainty (or model uncertainty) quantifies what the model does not know, which can be crucial for safe and efficient exploration, especially in unknown environments. A direction to explore for better uncertainty estimation is Bayesian learning based MARL.
\item \textit{Train using a larger Vicuna network or llama-2 when it’s publicly available} - Training with larger, more capable models such as a Vicuna-13B network or the Llama-2 could potentially yield significant improvements in planning capabilities of the oracle planner due to their increased model capacity and generalization ability. 
\item \textit{Fine tune using M* instead of using A* as M* is a multi-agent planning algorithm} - M* algorithm, being a multi-agent pathfinding solution, naturally fits into MARL setups where inter-agent coordination is crucial. By fine-tuning exploration strategies using M* instead of A*, one could enhance the collaborative aspects of the agents' behaviors, enabling more efficient multi-agent planning.
\item Apply this exploration strategy in different environments and different number of agents to understand the generalizability and scalability of the “expert” planner.
\item \textit{Apply this exploration strategy to a more complex action space with actions for tasks} - Extending the exploration strategy to handle more complex action spaces, where actions could involve multiple steps or hierarchical structures, can significantly broaden the applicability of this work. Such application could also be used to validate the generalizability of the approach in this work.

\end{enumerate}

\section{Source Code Overview}

The source code repository for this work has one main directory called \textit{llm\_marl}. This directory has been structured to facilitate easy navigation and understanding of the various components of the work. The repository is divided into several directories, each serving a specific function. Additionally, the \textit{llm\_marl} directory contains the Python modules. The sub-directories and the files are described below.

\begin{enumerate}
    \item \textit{config} - This subdirectory contains a python module named \textit{config\_loader.py} which is responsible for configuration files. It only support \textit{YAML}-based configuration files in this implementation. There are existing \textit{YAML} files for the experiments performed in this work. The configuration file contains trainer, experiment, model and algorithm-specific configurations and parameters.
    \item \textit{logger} - This subdirectory contains the \textit{logger.py} Python module which provides a function for creating a logger object.
    \item \textit{utils} - This subdirectory contains a Python module named `utils.py` which defines a class called `Utils` that contains several static methods for various utility functions. Here is a brief summary of some of the static method:
    \begin{enumerate}
        \item \textit{load\_config(config\_path, obj)}: This method loads a configuration file specified by \textit{config\_path} and sets the corresponding attributes of the \textit{obj} object.
        \item \textit{create\_sars(observation, next\_observation, action, reward, termination, truncation, device)}: This method creates a dictionary \textit{(sars\_dict)} that represents a state-action-reward-next-state tuple. The method converts the input data into Torch tensors and expands the dimensions as necessary.
        \item \textit{write\_meta\_data(base\_dir, obj)}: This method writes metadata information to a file specified by base\_dir. The metadata includes information about the environment, agent, experiment name, model configuration, algorithm specific information, etc. The method checks if the obj object has the corresponding attributes before writing them to the file.
        \item \textit{make\_experiment\_entry(base\_dir, experiment\_name)}: This method appends an entry to an "experiments.txt" file located in the parent directory of base\_dir. The entry includes the experiment name, current timestamp, and the base directory.
    \end{enumerate}
    \item \textit{ma\_models.py} - This file contains various classes that implements the critic for multi agent reinforcement learning, viz., \textit{BaseMultiAgentMLP} and \textit{DiscreteLinearMultiAgentCritic}. The \textit{BaseMultiAgentMLP} contains the base network architecture for an MARL actor or critic network and the \textit{DiscreteLinearMultiAgentCritic} uses this base network to implement a critic which is inturn used by the mixer network described below.
    \item \textit{mixer.py} - This file contains the \textit{QMixingNetwork} class which implement the mixer network as described in \cite{rashid2018qmix}.
    \item \textit{multi\_agent\_trainer.py} - This file defines the \textit{BaseMultiAgentTrainer} class that contains declarations of abstract method for QMIX training and contains an implementation of trajectory collection methods. The \textit{BaseMultiAgentTrainer} is inherited by the concrete class described below.
    \item \textit{multi\_agent\_off\_policy\_trainer.py} - This file defined a class named \textit{OffPolicyTrainerIndependentAgents} that contains the implementation of the \textit{QMIX} training loop and the agent evaluation. 
    \item \textit{multi\_agent\_qmix\_discrete\_env.py} - This file defines a class called \textit{MultiAgentQMIXExperiment} that initialized the objects required a training experiment for the \textit{QMIX} algorithm. The \textit{start} method is responsible for starting the training loop for \textit{QMIX}. 
    \item \textit{qmix.py} - The code in this file implements the QMIX algorithm for multi-agent reinforcement learning. It includes classes for A* search oracle, QMixingNetwork, and Utils. The QMIX class initializes the QMixingNetwork and performs updates using a replay buffer. It also includes methods for applying Vicuna-7B as a planner and fine-tuning a Vicuna-7B model and generating actions based on observations. 
\end{enumerate}

The source code for this work is available here: https://github.com/ashishkumar88/large-language-models-project.
\bibliography{references}

\end{document}